\pdfoutput=1

\documentclass[11pt]{article}

\usepackage[]{acl}

\usepackage{times}
\usepackage{latexsym}

\usepackage[T1]{fontenc}
\usepackage[utf8]{inputenc}

\usepackage{subcaption}
\usepackage{microtype}

\usepackage{multirow}
\usepackage{graphicx}
\usepackage{CJKutf8}
\usepackage{mathtools}
\DeclareMathAlphabet\mathbfcal{OMS}{cmsy}{b}{n}
\title{{\fontsize{13.91pt}{1pt}\selectfont  Pretraining Chinese BERT for Detecting Word Insertion and Deletion Errors}}

\author{Cong Zhou$^1$\thanks{\ \ \ Work done during internship at Tencent AI Lab.}, \ \ Yong Dai$^2$, Duyu Tang$^2$\thanks{ \ \  Contact: Duyu Tang ({duyutang@tencent.com}) }\ , Enbo Zhao$^2$,\\ \textbf{ Zhangyin Feng$^2$, Li Kuang$^1$, \and Shuming Shi$^2$} \\ \\   
	$^1$ Central South University \\ $^2$ Tencent AI Lab\\}

\begin{document}
\maketitle
\begin{CJK*}{UTF8}{gbsn}
\begin{abstract}
Chinese BERT models achieve remarkable progress in dealing with grammatical errors of word substitution. 
However, they fail to handle word insertion and deletion because BERT assumes the existence of a word at each position.
To address this, we present a simple and effective Chinese pretrained model.
The basic idea is to enable the model to determine whether a word exists at a particular position.
We achieve this by introducing a special token \texttt{[null]}, the prediction of which stands for the non-existence of a word.
In the training stage, we design pretraining tasks such that the model learns to predict \texttt{[null]} and real words jointly given the surrounding context. 
In the inference stage, the model readily detects whether a word should be inserted or deleted with the standard masked language modeling function.
We further create an evaluation dataset to foster research on word insertion and deletion. 
It includes human-annotated corrections for 7,726 erroneous sentences.
Results show that existing Chinese BERT performs poorly on detecting insertion and deletion errors.
Our approach significantly improves the F1 scores from 24.1\% to 78.1\% for word insertion and from 26.5\% to 68.5\% for word deletion, respectively.
\end{abstract}

\section{Introduction}
Grammatical Error Correction (GEC), which aims to detect and correct grammatical errors of text, is an important and active research area in natural language processing \cite{ng2014conll,napoles2017jfleg,ge2018reaching,awasthi2019parallel,omelianchuk2020gector,rothe2021simple}. 
In this work, we study Chinese GEC \cite{chang1995new,huang2000automatic,yu-li-2014-chinese,yu2014overview,zhang-etal-2015-hanspeller,li-etal-2021-exploration}, which has unique challenges as Chinese language does not have word delimiters (i.e., spaces) in written sentences.
Major types of grammatical errors include word\footnote{For simplicity, we use the term ``word'' to represent a Chinese token that may be composed of one or more characters.} substitution, insertion and deletion. Recently, BERT and its model variations \cite{hong2019faspell,zhang-etal-2020-spelling,liu-etal-2021-plome} show promising results on handling grammatical errors related to word substitution. 
However, they fail to handle word insertion and deletion errors, both of which are crucial and the sum of these two types account for 44.7\% of the grammatical errors in the CGED dataset \cite{rao-etal-2020-overview}.
The reason why it can't be handled well is that the learning objective of BERT assumes the existence of a word at each position, thus it is
incapable of determining whether no word exists at a position. 
Take word insertion as an example. If a word needs to be inserted between two words (i.e., $w_i$ and $w_{i+1}$), the standard BERT hardly detects anomalies because $w_i$ fits well to its preceding context and $w_{i+1}$ fits well to its following context.

We present a new Chinese pretrained model to address the aforementioned issue.
Our model inherits from BERT the power of contextual word prediction, and is further 
enhanced with the ability to determine whether no word should be predicted.
We achieve this by introducing a special token \texttt{[null]}, the prediction of which represents the non-existence of a word.
Our model is trained with masked language modeling (MLM), where a \texttt{[mask]} token can be 
inserted between two input words or substituted from an input word.
For the former, the pretraining task is to predict \texttt{[null]}.
For the latter, the pretraining task falls back to the original MLM objective of BERT.
Since our model is a tailored BERT, we initialize both Transformer blocks and the head layer at top with a public Chinese BERT. 
The training of our model converges fast in practice. 
Our model can be easily adopted to handle word insertion and deletion with the standard MLM function.
When examining whether a word needs to be inserted between $w_i$ and $w_{i+1}$, we insert a \texttt{[mask]} token between them
and check if the probability of \texttt{[null]} being predicted at the position of \texttt{[mask]} is low (e.g., lower than a threshold of 10.0\%).
An additional advantage of the model is that it conducts detection and correction simultaneously because the top ranked prediction can be directly used for correction.
When detecting whether a word $w_i$ needs to be deleted, we check if the probability of \texttt{[null]} being predicted at the position of $w_i$ is high (e.g., higher than a threshold of 99.0\%). 

We further contribute by creating an evaluation set of human-annotated sentences.
It consists of 4,969 and 2,757 sentences\footnote{The annotation process is still ongoing, so that we could expect an evaluation set with more annotated examples later. } with annotated corrections for insertion and deletion errors, respectively. 
We find that the standard Chinese BERT performs poorly on 
the detection of insertion and deletion errors.
On detecting insertion errors, our approach improves the F1 score from 24.1\% to 78.1\%.
On detecting deletion errors, our approach improves the F1 score from 26.5\% to 68.5\%.

\section{Background on MLM and Notation}\label{sec:method-mlm}
To make the paper self-contained, we briefly describe the standard  learning objective of MLM in this subsection. 
The basic idea of MLM is to reconstruct the corrupted words of a sequence given the surrounding context. 
Specifically, given an original sequence $x = (x_1, x_2, ... , x_n)$ consisting of $n$ words, a part of words are ``corrupted'' through being substituted by special \texttt{[mask]} tokens or replaced by randomly sampled words from a vocabulary of tens of thousands words.
Let's denote the corrupted input sequence as  $\tilde{x}=(\tilde{x}_1, \tilde{x}_2, ... , \tilde{x}_m)$, which is composed of $m$ words, and denote the set of the indexes of corrupted words as $I(\tilde{x})$.
It is worth noting that $\tilde{x}$ and $x$ have the same amount of words (i.e., $m = n$) and the words whose indexes do not fall into $I(\tilde{x})$ are identical in $\tilde{x}$ and $x$.

The goal of MLM is to produce $x$ given $\tilde{x}$. 
This is typically formulated as a cloze task, which only reconstructs the corrupted words
given context. The loss function is formulated as follows, 
\begin{align}
    \mathcal{L_\text{MLM}} = - \sum_{i \in I(\tilde{x})} \text{log} \, p(y_i= t_i | h(\tilde{x})_i)
\end{align}
where $h(\tilde{x})_i$ is the contextual representation of the $i$-th word in $\tilde{x} $ calculated by Transformer \cite{vaswani2017attention} and $p(y_i = t_i | h(\tilde{x})_i)$ is the conditional probability of the ground truth $t_i$ to be predicted. In the standard BERT, $t_i$ is equals to $x_i$.
In the implementation of BERT, corrupted words account for 15\% tokens of the input sequence. A corrupted word can be substituted with a \texttt{[mask]} token (80\% of the time), substituted with a random word (10\% of the time), and unchanged (10\% of the time).

\section{Method}
In this section, we first introduce our pretraining task. Then, we present the application of our model to word insertion and deletion, respectively.

\begin{figure*}[h]
	\centering
	\includegraphics[width=\textwidth]{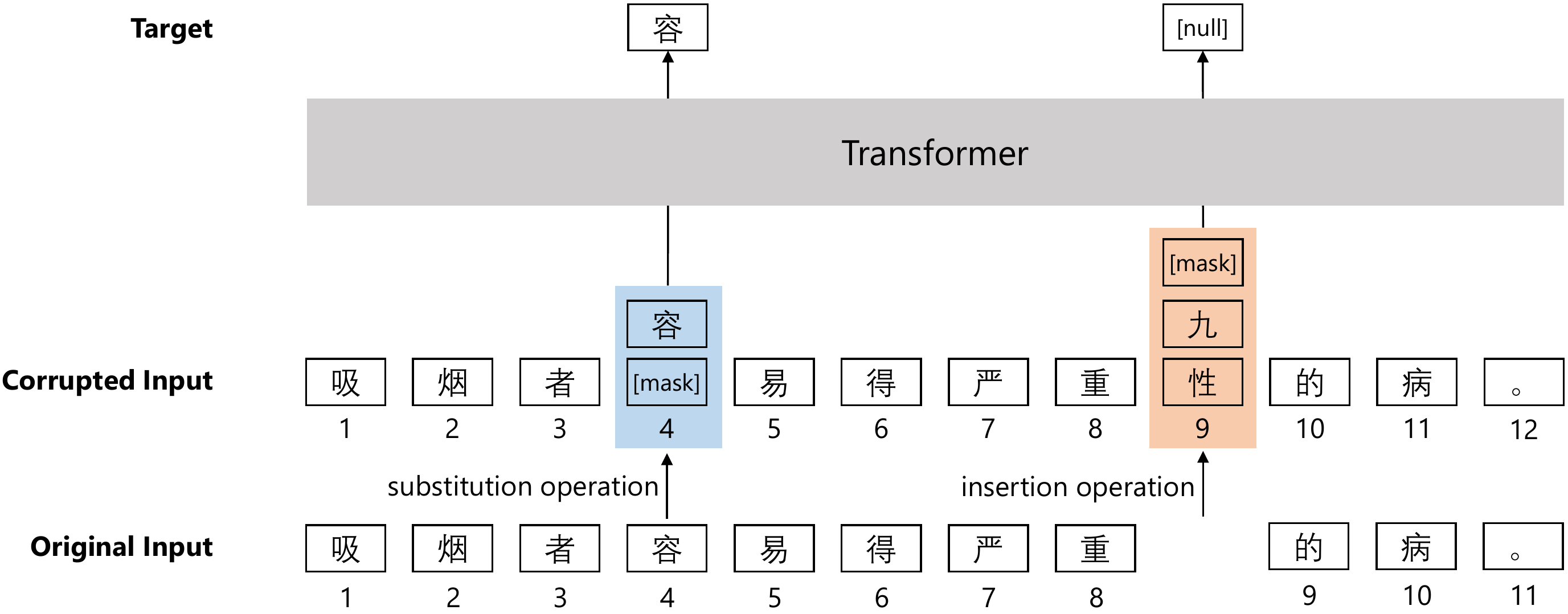}
	\caption{An example of our data corruption process during training. The original sentence is ``吸烟者容易得严重的病。'' (translation in English: ``Smokers are prone to serious illness''). Substitute operation is applied to the fourth word. We have two options (marked in blue) through replacing the word with \texttt{[mask]} or keeping the word unchanged. Insert operation is adopted between the 8th and 9th words. There are three options (marked in orange), including \texttt{[mask]}, a randomly selected word, and a word produced with another Chinese BERT model.}
	\label{fig:train-fig}
\end{figure*}

\subsection{Learning to Predict \texttt{[null]}}\label{sec:our-pretraining-task}
We introduce our pre-training task in this subsection.
The basic idea is that our model inherits from BERT the ability to select the most suitable word from the vocabulary given context, and is further enhanced with the ability to detect whether no word should occur at a particular position. 

We achieve this by introducing a special token \texttt{[null]}, the prediction of which stands for the non-existence of a word.
Different from BERT that corrupts the input only with word substitution, we produce a corrupted sequence with both word insertion and substitution operations.
Consider an original input $x = (x_1, x_2, ... , x_n)$ consisting of $n$ words.
If word insertion is applied to word $x_i$, we keep $x_i$ unchanged and insert a word $\tilde{x}^{ins}_i$ after $x_i$ (details about $\tilde{x}^{ins}_i$ are introduced next). 
As a result, the corrupted sequence $\tilde{x}$ includes a subsequence of ($x_i$, $\tilde{x}^{ins}_i$, $x_{i+1}$)
and the length of $\tilde{x}$ is no less than the length of $x$ (i.e., $m \geq n$).
In the training stage, the goal is predicting \texttt{[null]} at the position of $\tilde{x}^{ins}_i$, which means that there is no word between $x_i$ and $x_{i+1}$.
If word substitution is applied to word $x_i$, we largely follow the training process of BERT to replace $x_i$ with \texttt{[mask]} or keep $x_i$ unchanged.

The overall loss function is given as follows, where $I_{sub}(\tilde{x})$ and $I_{ins}(\tilde{x})$ are the indexes of corrupted words produced with substitution and insertion operations, respectively. 
\begin{align}\label{equ-our-loss}
    \begin{split}
    \mathcal{L_\text{ours}} = 
    & - \sum_{i \in I_{sub}(\tilde{x})} \text{log} \, p(y_i= t_i | h(\tilde{x})_i)\\
    & - \sum_{i \in I_{ins}(\tilde{x})} \text{log} \, p(y_i= \texttt{[null]} | h(\tilde{x})_i)\\
    \end{split}
\end{align}
The former part of the loss helps the model inherit the ability of contextual word prediction from BERT, and the latter part helps to learn whether a word should exist. 

An illustration of the data corruption process is depicted in Figure \ref{fig:train-fig}. We perform the following strategies to corrupt original texts:

$\bullet$ $\mathbfcal{R}$\textbf{1}. We randomly sample 15\% of input words for corruption, among which half of them are corrupted with the substitution operation and the other half are obtained with the insertion operation.

$\bullet$ $\mathbfcal{R}$\textbf{2}. For the insertion operation, the newly inserted word $x^{ins}_i$ is \texttt{[mask]} 50\% of the time and is randomly selected from the vocabulary 15\% of the time. 
For the remaining 35\%, we use a mask-and-generate pipeline to produce a real word as $x^{ins}_i$.
Specifically, we first insert a \texttt{[mask]} token at the position of $x^{ins}_i$, and then adopt MLM of a standard BERT model \cite{devlin2018bert} to produce top 10 words. Finally, we randomly select one of them as $x^{ins}_i$. This is effective for handling word deletion (see results at Section \ref{sec:results-mask-and-generate}).

$\bullet$ $\mathbfcal{R}$\textbf{3}. For the substitution operation, a word has a fifty-fifty chance of being replaced with \texttt{[mask]} or keeping unchanged.

\subsection{Application to Word Insertion}
\label{3.3}
\begin{figure}[h]
	\centering
	\includegraphics[width=0.49\textwidth]{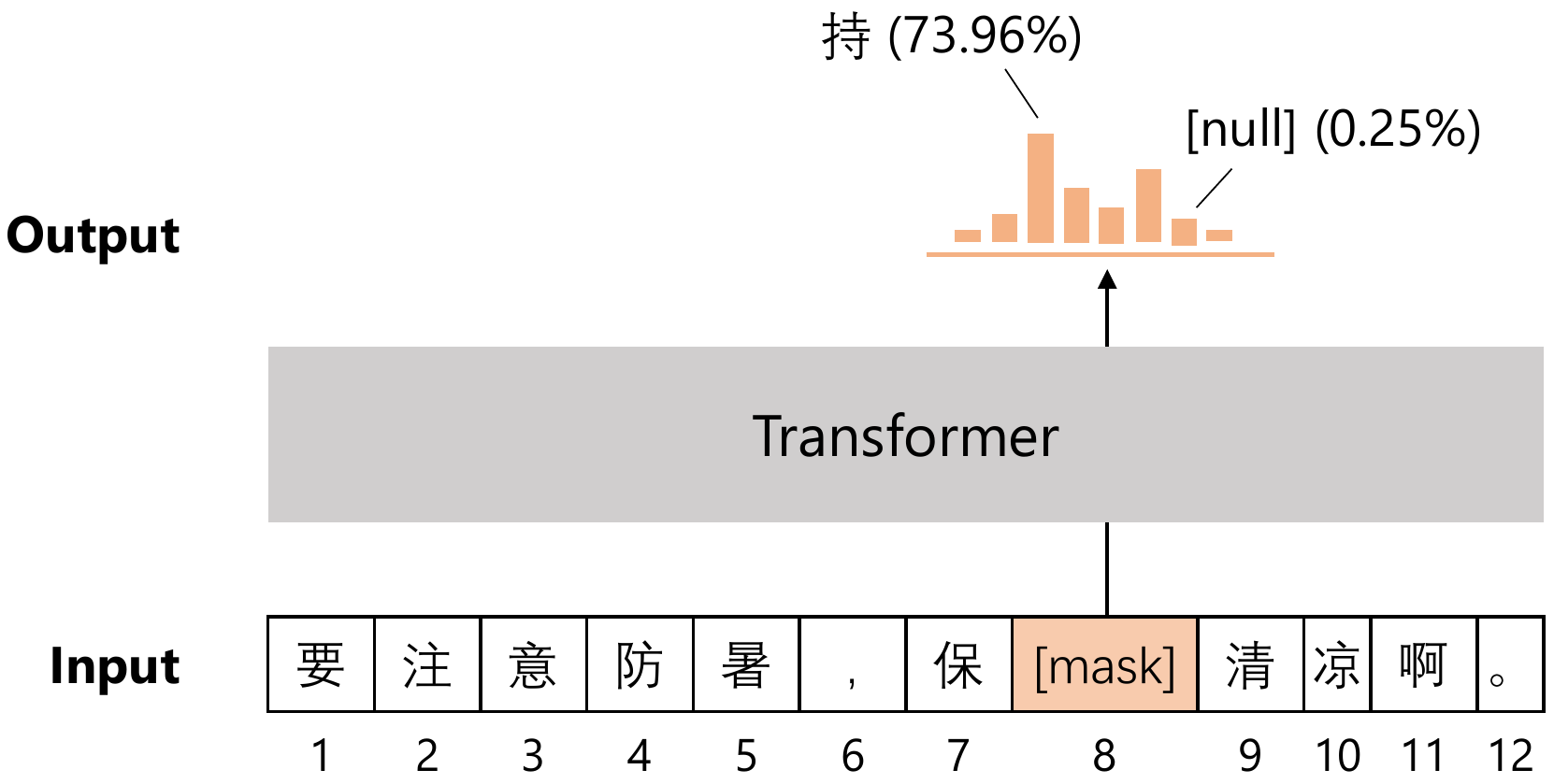}  
	\caption{An example of the inference process for word insertion. The translation is ``Prevent heatstroke and keep calm.''. When determining whether a word should be inserted between ``保'' and ``清'', we insert a \texttt{[mask]} token between them and check the probability of \texttt{[null]} with MLM. 
}
	\label{fig:inference-insert}
\end{figure}

In this subsection, we present how to apply our model to handling the grammatical error of word insertion. 
Given a sentence as the input, the task is to 
detect whether a word needs to be inserted between any two words and if so, generate the word to be inserted. 
We divide the whole process into two stages: detection and correction.
We use one model to accomplish two stages. An illustration is given in Figure \ref{fig:inference-insert}.

The goal of detection is to predict whether or not a word should be inserted between two words. 
In the inference phase, we first insert a \texttt{[mask]} token between $x_i$ and $x_{i+1}$, resulting in a new sequence of $n+1$ words.
Afterwards, we use our model to compute the probability of \texttt{[null]} to be predicted. If the probability is lower than a threshold (e.g., 10\%), a word needs to be inserted here. 
An advantage of the model is that the correction result can be obtained at the same time.  We regard the word with the highest probability as the word to be inserted.

\subsection{Application to Word Deletion}
We present how to apply our model to dealing with grammatical errors of word deletion. 
Given a sentence, the task is to detect whether a word should be deleted or not.
Compared to word insertion, correction is not needed for this task.

\begin{figure}[h]
	\centering
	\includegraphics[width=0.48\textwidth]{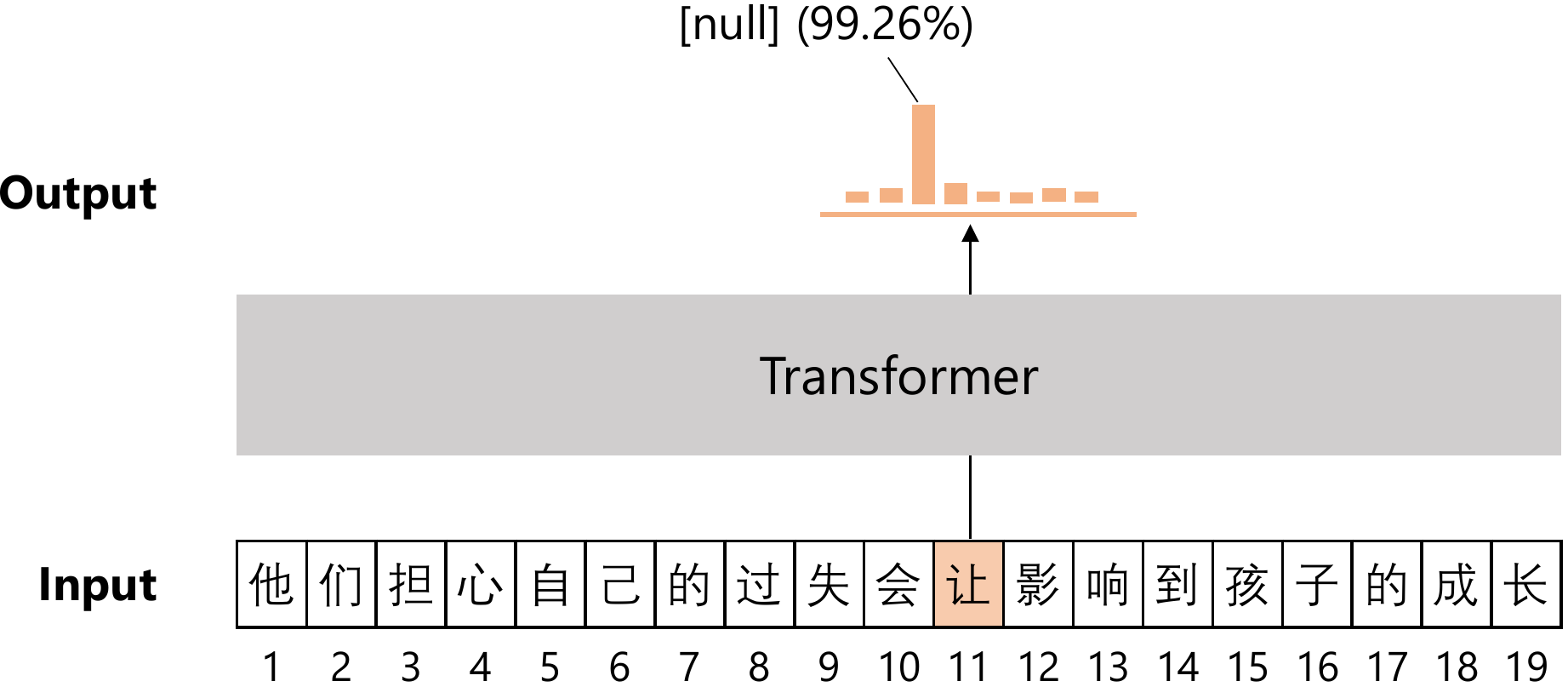}
	\caption{An example of the inference process for word deletion. The translation is ``They worry that their mistakes will affect the growth of their children.''. We directly conduct MLM without masking words and check the probability of \texttt{[null]} for each position.}
	\label{fig:inference-delete}
\end{figure}

Unlike the use of our model in word insertion, we do not insert \texttt{[mask]} tokens to the input. 
Here, we directly take the original sentence as the input and make predictions on top of the contextual representation of each word.
This is efficient in practice because detecting all tokens can be done in one forward pass.
Specifically, we check the probability of \texttt{[null]} being predicted for each word. 
If the probability is higher than a threshold (e.g., 99\%), the word should be deleted. 
Figure \ref{fig:inference-delete} shows an example of the inference process. 
For reduplication errors (e.g. ``的的''), we delete the last word.

\section{Dataset Annotation}

Existing datasets for Chinese grammatical error correction are either composed of sentences written by non-naive speakers of Chinese language \cite{zhao2018overview,rao-etal-2020-overview} or dominated by errors of word substitution \cite{tseng2015introduction,wang2018hybrid}. 
Our new dataset differs from the previous ones in that the sentences in our dataset are written by native Chinese speakers and contain a lot of insertion and deletion errors. Therefore our dataset is more suitable for studying the problem of detecting insertion and deletion errors. In this section, we describe how our dataset is built.

\begin{figure*}[h]
	\centering
	\includegraphics[width=\textwidth]{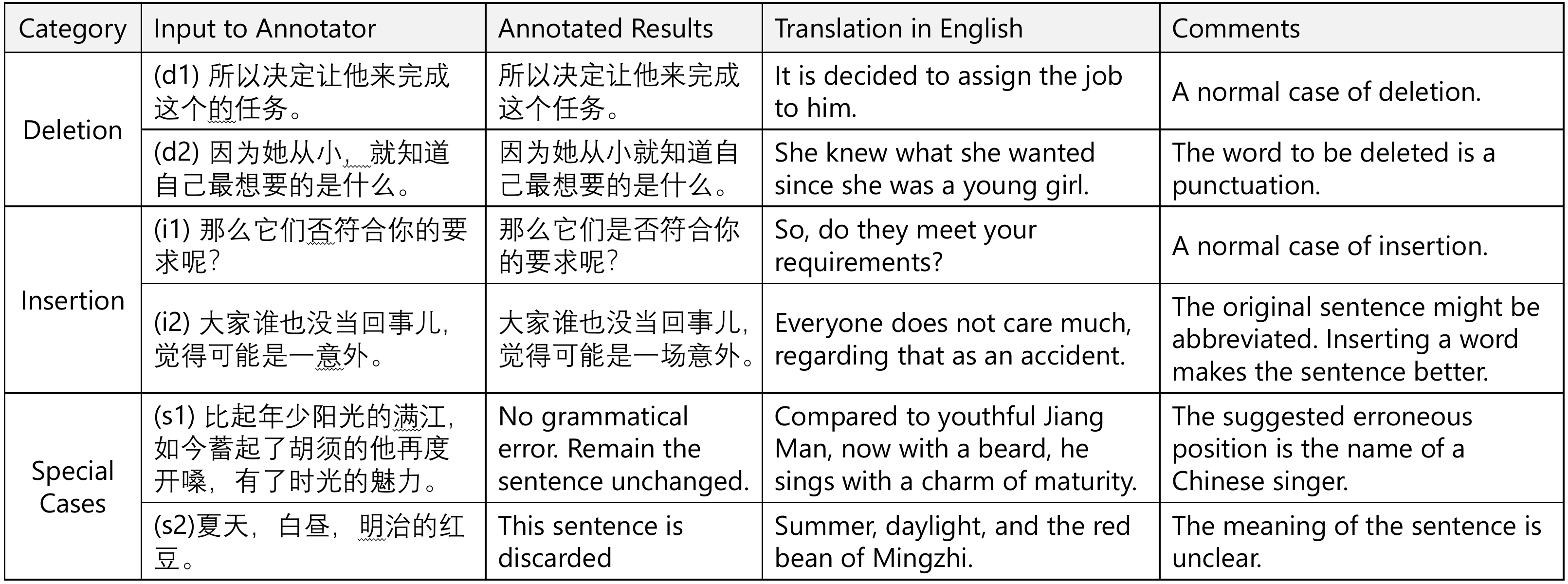}
	\caption{Examples of the annotated dataset. Each row stands for an instance sent to annotators. The word with wave underline in the 2nd column is the warning position which is also given to annotators. }
	\label{fig:annotate-example}
\end{figure*}

\subsection{Annotation Process}

An intuitive annotation pipeline is to first randomly sample sentences from the data source, and then dispatch them to human annotators to correct sentences with grammatical errors.
However, this is impractical because the majority of randomly sampled sentences do not contain grammatical errors.
To improve the annotation efficiency and make the experimental results convincing, we train a dedicated BERT model to select sentences for annotation. 
We train this model by corrupting a sentence with an inserted \texttt{[mask]} and is trained to predict \texttt{[null]}. Thus, this model is capable of producing warning positions (i.e., where the errors may occur).
We apply this model to web news and blog articles of many Gigabytes, resulting in 
sentences with the probabilities of having insertion/deletion errors as well as the warning positions.
We rank these sentence with the probabilities in descending order and send each sentence together with the warning position to 
at least one human annotator and one judge.
Annotators are asked to 
rewrite the sentence if there are grammatical errors. 
Annotators are reminded to avoid making huge changes to the sentence structure and are allowed to ignore some warning positions.
An instance is discarded if any of the following three situations is satisfied: (1) the judge does not agree on the annotated results;  (2) the meaning of the sentence is confusing; (3) there are no insertion or deletion errors. 

\subsection{Annotation Results}

Figure \ref{fig:annotate-example} gives examples of insertion and deletion types as well as some special cases. 
We can see that (d1) is a normal case of word deletion and (d2) shows a special case where a comma is deleted.
A general case of insertion is given in (i1). 
Sometimes, people may use abbreviated expressions in written sentences (e.g., i2), our criterion is that annotators are encouraged to insert a word if carry out insertion leads to a better sentence.
We also given two special cases. In (s1), the suggested position is a part of an named entity, so that  no action needs to be taken. In (s2), the whole sentence is discarded because the meaning of the sentence is unclear to both annotators and judges. 
From a randomly sampled set of discarded sentences, we find that most of them are written in traditional Chinese, are largely composed of ancient Chinese poems, or include many colloquial and slang words.

After the annotation process is completed, we implement a text alignment toolkit for Chinese language on top of ERRANT \citep{bryant-etal-2017-automatic}\footnote{\url{https://github.com/chrisjbryant/errant}}, a commonly used toolkit to automatically annotate parallel English sentences consisting of erroneous and corrected sentence pairs.
We use our tool to obtain the positions of errors and the corrections through aligning the original sentences and the annotations. 
A special case is reduplication error (e.g.  “的的”). If the annotation result is to delete one of the duplicated words, we mark the last word as the erroneous position.
Table \ref{lab:data-statistic} shows the statistics of our data consisting of insertion and deletion types of errors.
The dataset will be made publicly available to the community.

\begin{table}[h]\small
\centering
\begin{tabular}{l|c|c}
\hline
Dataset              & Deletion     & Insertion  \\ \hline
Count           & 2,757        & 4,969         \\  \hline
No. words per sentence         & 59.4        & 55.6         \\
Erroneous positions per sentence  & 1.06         & 1.24          \\ 
No. sentence w/ one error   & 2,612 & 3,900 \\
No. sentence w/ two errors  & 116   & 874  \\
No. sentence w/ three errors & 15    & 153    \\ \hline
\end{tabular}
\caption{Statistics of the annotated dataset.}
\label{lab:data-statistic}
\end{table}

\section{Experiments}
We report results and model analysis in this section.
\subsection{Setting}
We evaluate on our annotated dataset.
For each insertion and deletion type, we keep a subset of 500 instances as the development set and leave the remaining as the test set. 
We don't conduct experiments in a supervised setting, so no data points remain for fine-tuning. The development set is only used for selecting a threshold.  Evaluation results are reported on the test set.
The model can  be easily used in a supervised setting, where a binary classifier is learned on top of the representation of \texttt{[mask]}, we leave this as a future work. 

We evaluate in terms of two subtasks: error detection and error correction.
Detection is required for both word insertion and deletion. Following previous work \cite{wang2019confusionset,cheng-etal-2020-spellgcn,liu-etal-2021-plome}, we use character-level precision, recall, and F1 scores as the evaluation metrics. 
We calculate these metrics based on the script of \citet{wang2019confusionset}\footnote{\url{https://github.com/wdimmy/Automatic-Corpus-Generation}}.
Correction is only applicable to word insertion. We evaluate in an end-to-end manner: an output is correct only if both the detected position and the revised word are correct. 

Our model largely follows BERT-base \cite{devlin2018bert}. It has 12 Transformer encoder blocks, 12 self-attention heads and 768 hidden state dimensions.
We increase the vocabulary size of BERT by one (i.e., adding a special token of \texttt{[null]}).
The parameters of the embedding layer, the Transformer blocks, and the head layer except for parameters related to \texttt{[null]} are initialized from an existing Chinese BERT \cite{devlin2018bert}\footnote{\url{https://github.com/google-research/bert}}.
We collect web news and blog articles as the pre-training data, which has about 800 Gigabytes after data processing. 
We set batch size as 10,240 and train the model on 32 Tesla V100 GPUs. 
We set the learning rate as 1e-4 and use the Adam optimizer \cite{kingma2014adam}  with a linear warmup scheduler.
We find that the model converges fast after about 4K training steps.

\begin{table*}[h]
\centering
\begin{tabular}{l|ccc|ccc}
\hline
\multicolumn{1}{c|}{\multirow{2}{*}{Method}} & \multicolumn{3}{c|}{Insertion}                 & \multicolumn{3}{c}{Deletion} \\ \cline{2-7} 
\multicolumn{1}{c|}{}                        & Precision & Recall & F1   & Precision   & Recall  & F1    \\ \hline
\multicolumn{1}{l|}{BERT w/ substituted \texttt{[mask]}}               & 13.1      & 22.7   & \multicolumn{1}{l|}{16.5} & 17.1        & 21.0    & 18.8  \\
\multicolumn{1}{l|}{BERT w/o \texttt{[mask]}}            & 14.3      & 16.5    & \multicolumn{1}{l|}{15.3} & 23.0       & 31.3    & 26.5  \\
\multicolumn{1}{l|}{BERT w/ inserted \texttt{[mask]}}             & 28.0          & 21.2       & \multicolumn{1}{l|}{24.1}     &  -           &   -      &    -   \\
\multicolumn{1}{l|}{MacBERT}  & 6.4          &  21.2      & \multicolumn{1}{c|}{9.8}     & 10.8           &  33.5       &  16.3     \\ 
\multicolumn{1}{l|}{PLOME}                   & 6.7          & 26.5       & \multicolumn{1}{l|}{10.7}     & 10.6           &  35.2       &  16.3     \\ 
\multicolumn{1}{l|}{Flying}                   & 62.8          & 19.2       & \multicolumn{1}{l|}{29.4}     & 56.2           &  29.9       &  39.1     \\ 
\multicolumn{1}{l|}{Our approach}              & \textbf{77.6}      & \textbf{78.6}   & \multicolumn{1}{l|}{\textbf{78.1}} & \textbf{71.4}        & \textbf{65.9}    & \textbf{68.5}  \\ \hline
\end{tabular}
\caption{Results of grammatical error detection for word insertion and word deletion.}
\label{table:results-detection}
\end{table*}

\subsection{Baseline Models}\label{section:baselines}
We implement the following baselines for model comparison. For fair comparison, we continue pretraining a stronger Chinese BERT model (also initialized with the checkpoint of \citet{devlin2018bert}) with the the same dataset used to train our approach, and use that to build the first three baselines given below\footnote{Our further pretrained Chinese BERT performs better than the checkpoint of \citet{devlin2018bert} (e.g., with an improvement of 2.4 F1 score on word detection).}.

$\bullet$ \textbf{BERT w/ substituted \texttt{[mask]}}. 
When we detect whether a word is erroneous, we replace it with \texttt{[mask]} and check whether the probability of the original word being predicted by MLM is lower than a threshold. 
For correction, we insert a \texttt{[mask]} token when the detection model decides to insert a word.
We run MLM with the same BERT model and output the word with the highest probability.

$\bullet$ \textbf{BERT w/o \texttt{[mask]}}. Compared to the first baseline, the only difference is that this baseline does not use \texttt{[mask]} token. The original word remains the same in the detection phase. This baseline is not applicable to correction. We also implement a baseline with the same pipeline using MacBERT \cite{cui2020revisiting}.

$\bullet$ \textbf{BERT w/ inserted \texttt{[mask]}}. We build this baseline for handling insertion. 
In the detection stage, we insert a \texttt{[mask]} token between two words and check if the probability of the top predicted word is lower than a threshold. 
In the correction stage, the top predicted word for the inserted \texttt{[mask]} token is returned. 

$\bullet$ \textbf{PLOME} \cite{liu-etal-2021-plome} is the first BERT-style pretrained model for Chinese spelling correction. It is trained by corrupting the input with phonologically and visually similar characters instead of \texttt{[mask]} tokens. 
We develop baselines based on their pipeline\footnote{\url{https://github.com/liushulinle/PLOME}} and tune thresholds on the development sets to detect for insertion and deletion. Since the training of PLOME does not use \texttt{[mask]}, we don't apply it to correction. 

$\bullet$ \textbf{Flying} \cite{wang-etal-2020-combining-resnet} is the leading system in the challenge of Chinese Grammatical Error Diagnosis (CGED). It detects error types of words via sequence labeling. The feature of each word is computed based on the word embedding and the output of Transformers. 
We collect sentences with insertion and deletion errors from CGED, and use them to train two baseline systems for detecting insertion and deletion errors, respectively.\footnote{We find that using datasets of both types to train a joint detector performs worse (by about 5\% F1 score) than training two task-specific detectors.}

\subsection{Results on Error Detection}
Table \ref{table:results-detection} shows the results of error detection for both insertion and deletion types. 
Among these unsupervised baselines, inserting \texttt{[mask]} on top of BERT is the strongest for detecting insertion type of error.
This makes sense because the task is to detect whether a word should be added, and this baseline  gives the potentially real word a real position.
For the deletion error type, the strongest unsupervised baseline is to directly make the prediction without masking words.
It is not surprising that PLOME does not perform well on dealing with insertion and deletion errors because it is optimized for the task of word substitution. 
Flying system is the only supervised baseline which is trained on a public dataset. We can see that it has a very low recall because the style of the training data is different from our dataset.
Overall, our approach consistently outperforms baselines by large margins. 

We show the performance of our approach on error detection with different thresholds in Appendix.
For each task, we select the threshold that obtains the best F1 score on the dev set, and use it for testing. 
Appendix gives the performance of all baselines with different thresholds.

\subsection{Effects of Corrupted Word Generation}\label{sec:results-mask-and-generate}
We show how the sampling strategy of mask-and-generate ($\mathbfcal{R}$\textbf{2} of insertion operation in Section \ref{sec:our-pretraining-task}) plays an important role in word deletion. 
It is worth to note that when our approach takes the original sentence as the input (instead of replacing words with \texttt{[mask]}) and detects all the words of the input sentence at once. 
This strategy has a big advantage in terms of inference speed (see section \ref{sec:speed-analysis} below).
From Table \ref{sec:results-mask-and-generate}, we can see that mask-and-generate significantly improves the accuracy of word deletion. 
We believe the reason is that corrupting text via inserting a random word gives an easy training instance, whereas inserting a word via  mask-and-generate produces more difficult context-dependent instances.

\begin{table}[h]
\centering
\begin{tabular}{l|ccc}
\hline
Method & {P} & {R} & {F1} \\ \hline
Our approach & \textbf{71.4} & \textbf{65.9} & \textbf{68.5} \\ 
Our approach w/o mag  & 71.2 & 40.5 & 51.6 \\ \hline
\end{tabular}
\caption{Effects of the sampling strategy of mask-and-generate (mag) for word deletion.}
\label{table:results-mask-and-generate}
\end{table}

\subsection{Results on Error Correction}
We report the end-to-end performance of error correction in Table \ref{table:results-correction}. 
Since generating a word via MLM needs to insert a \texttt{[mask]} token, the baselines without using \texttt{[mask]} are not considered here.
Unsurprisingly, our approach performs better than baselines because our detection model shows huge gains over the baselines. 
The point is that our approach is capable of detecting and correcting an error with one model and one forward pass.

\begin{table}[h]\small
\centering
\begin{tabular}{l|ccc}
\hline
Method & {P} & {R} & {F1} \\ \hline
BERT w/ substituted \texttt{[mask]} & 8.4 & 14.8  & 10.8 \\
BERT w/ inserted \texttt{[mask]} & 26.6 & 20.1 & 22.9 \\
Our approach & \textbf{58.3} & \textbf{59.1} & \textbf{58.7} \\ \hline
\end{tabular}
\caption{Results of error correction for word insertion.}
\label{table:results-correction}
\end{table}

\begin{figure*}[h]
	\centering
	\includegraphics[width=\textwidth]{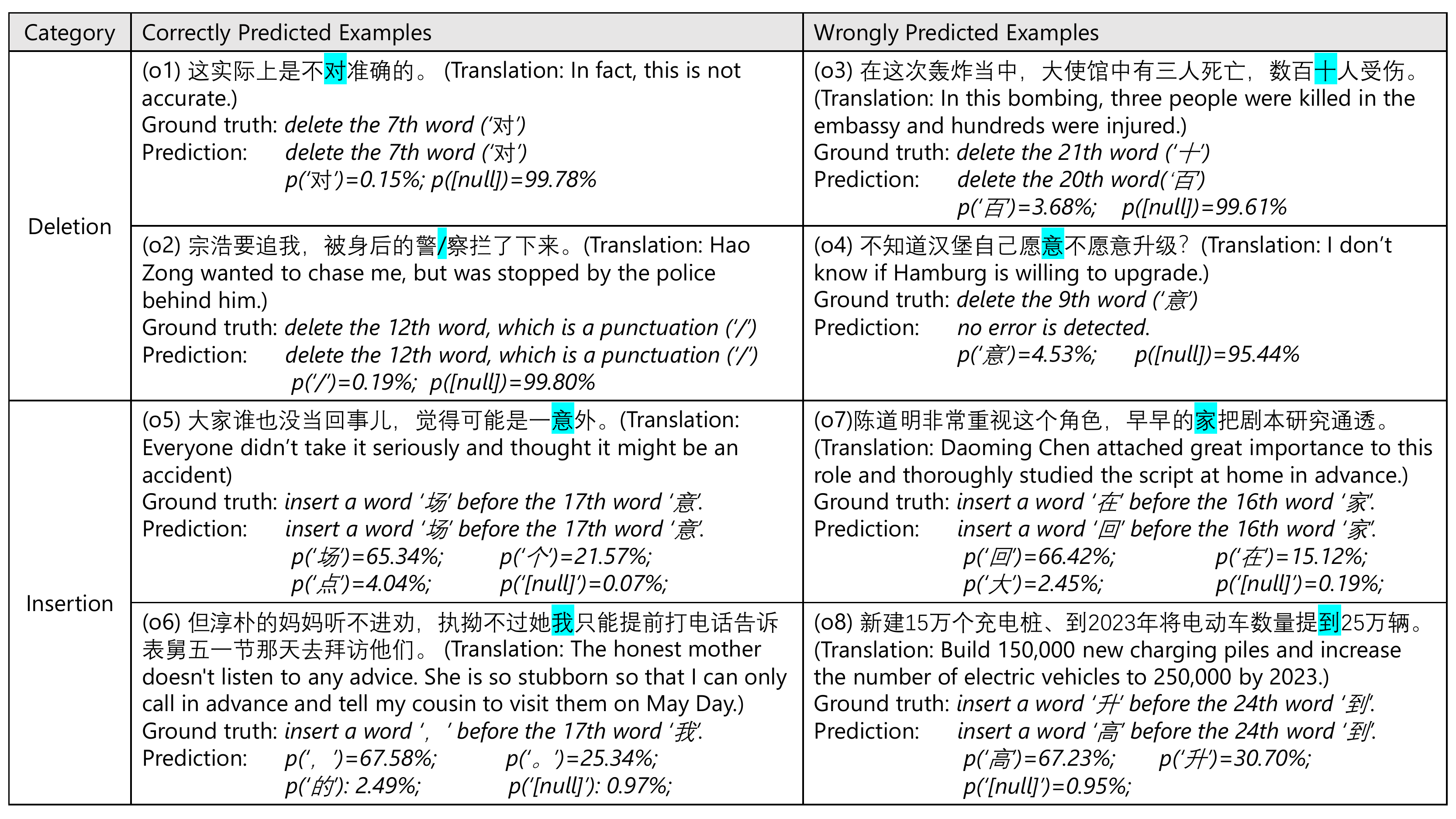}
	\caption{Correctly and wrongly predicted examples for word deletion and insertion. The word highlighted in blue in the 2nd column is where the error occurs. }
	\label{fig:case-study}
\end{figure*}

\subsection{Results on CGED Dataset}
To further demonstrate the broad effectiveness of our model, we conduct extended experiments on another widely adopted benchmark CGED~\cite{rao-etal-2020-overview}. Baseline systems described in Section \ref{section:baselines} are used.
We evaluate on the sentences containing insertion and deletion errors from CGED, respectively.
As Table \ref{table:results-cged} shows, our approach is obviously superior to baseline models in F1. Detailed results in P, R and F1 are given in the appendix.
\begin{table}[h]
\small
\centering
\begin{tabular}{l|c|c}
\hline
Method & {Insertion}                 &{Deletion} \\  \hline
\multicolumn{1}{l|}{BERT w/ substituted \texttt{[mask]}}              & \multicolumn{1}{c|}{1.5} & 6.2  \\
\multicolumn{1}{l|}{BERT w/o \texttt{[mask]}}    & \multicolumn{1}{c|}{12.2} & 28.2  \\
\multicolumn{1}{l|}{BERT w/ inserted \texttt{[mask]}}         & \multicolumn{1}{c|}{11.8}     &    -   \\
\multicolumn{1}{l|}{Flying}             & \multicolumn{1}{c|}{17.2}     &  20.8     \\ 
\multicolumn{1}{l|}{Our approach}            & \multicolumn{1}{c|}{\textbf{41.5}} & \textbf{40.4}  \\ \hline
\end{tabular}
\caption{Results on CGED in F1 score.}
\label{table:results-cged}
\end{table}

\subsection{Case Study and Error Analysis}
Figure \ref{fig:case-study} shows the correctly and wrongly predicted examples detected by our model. We can see that the model is capable of handling both normal words (o1 and o5) and punctuations (o2 and o6). 

In o3, our second-ranked prediction (``数百人'', hundreds of people) is same with the ground truth. 
Although the top prediction is incorrect, the result (``数十人'', dozens of people)
also fits well to the the context. 
In o4, the probability of \texttt{[null]} is as high as 95.44\%. However, our model does not detect any errors because our threshold is strict (i.e., 99.1\%).
In  o7 and o8, although both second-ranked predictions are same with the ground truth, the top predictions also make sense.

\subsection{Analysis on Inference Time Cost}\label{sec:speed-analysis}
We report the time costs of our approach in word deletion and insertion, respectively. 
For word deletion, our approach runs one forward pass to detect all the words of a sentence.
For word insertion, our approach runs one forward pass for each word, since detecting for a word (i.e., a position) needs to insert a \texttt{[mask]} token at the particular position and remaining other contexts untouched.
On average, the inference time costs for deletion and insertion are respectively 9.2ms and 106.3ms on one V100 GPU machine (the average sentence length is shown in Table \ref{lab:data-statistic}).
We attempted to speed up the inference speed for word insertion by inserting a \texttt{[mask]} between every two words, so that running one forward pass could handle all words. This improves the inference speed by almost 10 times, but with the cost of accuracy decrease (from 78.1 to 72.3 in terms of F1).

\section{Related Work}
We describe related works on Chinese GEC and pretrained models.

Recent methods for Chinese GEC can be generally categorized into sequence translation and sequence tagging.
Sequence translation methods  \citep{wang-etal-2018-hybrid,wang2019confusionset,wang-etal-2020-chinese,li2021tail} typically use sequence-to-sequence models based on LSTM or Transformer, where the encoder consumes grammatically erroneous sentences and the decoder generates correct sentences in an autoregressive or non-autoregressive manner.
Although these models are able to handle various error types, the generated results might be not faithful to the input because of the exposure bias issue \citep{zhang-etal-2019-bridging}.
Recent sequence tagging methods \citep{zhang-etal-2020-spelling,cheng-etal-2020-spellgcn,zhang2021correcting,xu2021read} 
usually use neural network architectures to first tag the error positions and then correct them. Among them, BERT-based models are capable of conducting detection and correction of substitution error at once, however,  
they fail to deal with insertion and deletion errors.

PLOME \citep{liu-etal-2021-plome} is the first pretrained model developed for Chinese grammatical error correction with a focus on handling word substitution. They use GRU to encode phonic and shape of Chinese character as additional input to the model.
Our experimental results indicate that PLOME does not adapt well to word insertion or deletion. It is an interesting future work to develop a versatile model that combines the advantages of PLOME and our approach.
There are general-purpose Chinese BERT models whose learning processing relates to GEC in some sense. For example, 
MacBERT \citep{cui2020revisiting} regards MLM as correction, where corrupted words are replaced with synonyms instead of \texttt{[MASK]} tokens.

\section{Conclusion}
This work proposes a pretrained model for dealing with grammatical errors of word insertion and deletion.
A special token \texttt{[null]} is introduced to facilitate the model to handily detect the existence of a word.
We further contribute by creating an evaluation dataset with human-labeled corrections.
Results show that existing BERT models perform poorly on detection the insertion and deletion of words. Our approach obtains significant gains on both tasks. 
In the future, we plan to develop a versatile model with the ability to handle many other types of grammatical errors.

\end{CJK*}
\bibliography{anthology,custom}

\begin{thebibliography}{33}
\expandafter\ifx\csname natexlab\endcsname\relax\def\natexlab#1{#1}\fi

\bibitem[{Awasthi et~al.(2019)Awasthi, Sarawagi, Goyal, Ghosh, and
  Piratla}]{awasthi2019parallel}
Abhijeet Awasthi, Sunita Sarawagi, Rasna Goyal, Sabyasachi Ghosh, and Vihari
  Piratla. 2019.
\newblock Parallel iterative edit models for local sequence transduction.
\newblock \emph{arXiv preprint arXiv:1910.02893}.

\bibitem[{Bryant et~al.(2017)Bryant, Felice, and
  Briscoe}]{bryant-etal-2017-automatic}
Christopher Bryant, Mariano Felice, and Ted Briscoe. 2017.
\newblock \href {https://doi.org/10.18653/v1/P17-1074} {Automatic annotation
  and evaluation of error types for grammatical error correction}.
\newblock In \emph{Proceedings of the 55th Annual Meeting of the Association
  for Computational Linguistics (Volume 1: Long Papers)}, pages 793--805,
  Vancouver, Canada. Association for Computational Linguistics.

\bibitem[{Chang(1995)}]{chang1995new}
Chao-Huang Chang. 1995.
\newblock A new approach for automatic chinese spelling correction.
\newblock In \emph{Proceedings of Natural Language Processing Pacific Rim
  Symposium}, volume~95, pages 278--283. Citeseer.

\bibitem[{Cheng et~al.(2020)Cheng, Xu, Chen, Jiang, Wang, Wang, Chu, and
  Qi}]{cheng-etal-2020-spellgcn}
Xingyi Cheng, Weidi Xu, Kunlong Chen, Shaohua Jiang, Feng Wang, Taifeng Wang,
  Wei Chu, and Yuan Qi. 2020.
\newblock \href {https://doi.org/10.18653/v1/2020.acl-main.81} {{S}pell{GCN}:
  Incorporating phonological and visual similarities into language models for
  {C}hinese spelling check}.
\newblock In \emph{Proceedings of the 58th Annual Meeting of the Association
  for Computational Linguistics}, pages 871--881, Online. Association for
  Computational Linguistics.

\bibitem[{Cui et~al.(2020)Cui, Che, Liu, Qin, Wang, and Hu}]{cui2020revisiting}
Yiming Cui, Wanxiang Che, Ting Liu, Bing Qin, Shijin Wang, and Guoping Hu.
  2020.
\newblock Revisiting pre-trained models for chinese natural language
  processing.
\newblock \emph{arXiv preprint arXiv:2004.13922}.

\bibitem[{Devlin et~al.(2018)Devlin, Chang, Lee, and
  Toutanova}]{devlin2018bert}
Jacob Devlin, Ming-Wei Chang, Kenton Lee, and Kristina Toutanova. 2018.
\newblock Bert: Pre-training of deep bidirectional transformers for language
  understanding.
\newblock \emph{arXiv preprint arXiv:1810.04805}.

\bibitem[{Ge et~al.(2018)Ge, Wei, and Zhou}]{ge2018reaching}
Tao Ge, Furu Wei, and Ming Zhou. 2018.
\newblock Reaching human-level performance in automatic grammatical error
  correction: An empirical study.
\newblock \emph{arXiv preprint arXiv:1807.01270}.

\bibitem[{Hong et~al.(2019)Hong, Yu, He, Liu, and Liu}]{hong2019faspell}
Yuzhong Hong, Xianguo Yu, Neng He, Nan Liu, and Junhui Liu. 2019.
\newblock Faspell: A fast, adaptable, simple, powerful chinese spell checker
  based on dae-decoder paradigm.
\newblock In \emph{Proceedings of the 5th Workshop on Noisy User-generated Text
  (W-NUT 2019)}, pages 160--169.

\bibitem[{Huang et~al.(2000)Huang, Pan, Ming, and Zhang}]{huang2000automatic}
Changning Huang, Haihua Pan, Zhou Ming, and Lei Zhang. 2000.
\newblock Automatic detecting/correcting errors in chinese text by an
  approximate word-matching algorithm.

\bibitem[{Kingma and Ba(2014)}]{kingma2014adam}
Diederik~P Kingma and Jimmy Ba. 2014.
\newblock Adam: A method for stochastic optimization.
\newblock \emph{arXiv preprint arXiv:1412.6980}.

\bibitem[{Li et~al.(2021)Li, Zhang, Zheng, and
  Huang}]{li-etal-2021-exploration}
Chong Li, Cenyuan Zhang, Xiaoqing Zheng, and Xuanjing Huang. 2021.
\newblock \href {https://doi.org/10.18653/v1/2021.acl-short.56} {Exploration
  and exploitation: Two ways to improve {C}hinese spelling correction models}.
\newblock In \emph{Proceedings of the 59th Annual Meeting of the Association
  for Computational Linguistics and the 11th International Joint Conference on
  Natural Language Processing (Volume 2: Short Papers)}, pages 441--446,
  Online. Association for Computational Linguistics.

\bibitem[{Li and Shi(2021)}]{li2021tail}
Piji Li and Shuming Shi. 2021.
\newblock Tail-to-tail non-autoregressive sequence prediction for chinese
  grammatical error correction.
\newblock \emph{arXiv preprint arXiv:2106.01609}.

\bibitem[{Liu et~al.(2021)Liu, Yang, Yue, Zhang, and
  Wang}]{liu-etal-2021-plome}
Shulin Liu, Tao Yang, Tianchi Yue, Feng Zhang, and Di~Wang. 2021.
\newblock \href {https://doi.org/10.18653/v1/2021.acl-long.233} {{PLOME}:
  Pre-training with misspelled knowledge for {C}hinese spelling correction}.
\newblock In \emph{Proceedings of the 59th Annual Meeting of the Association
  for Computational Linguistics and the 11th International Joint Conference on
  Natural Language Processing (Volume 1: Long Papers)}, pages 2991--3000,
  Online. Association for Computational Linguistics.

\bibitem[{Napoles et~al.(2017)Napoles, Sakaguchi, and
  Tetreault}]{napoles2017jfleg}
Courtney Napoles, Keisuke Sakaguchi, and Joel Tetreault. 2017.
\newblock Jfleg: A fluency corpus and benchmark for grammatical error
  correction.
\newblock \emph{arXiv preprint arXiv:1702.04066}.

\bibitem[{Ng et~al.(2014)Ng, Wu, Briscoe, Hadiwinoto, Susanto, and
  Bryant}]{ng2014conll}
Hwee~Tou Ng, Siew~Mei Wu, Ted Briscoe, Christian Hadiwinoto, Raymond~Hendy
  Susanto, and Christopher Bryant. 2014.
\newblock The conll-2014 shared task on grammatical error correction.
\newblock In \emph{Proceedings of the Eighteenth Conference on Computational
  Natural Language Learning: Shared Task}, pages 1--14.

\bibitem[{Omelianchuk et~al.(2020)Omelianchuk, Atrasevych, Chernodub, and
  Skurzhanskyi}]{omelianchuk2020gector}
Kostiantyn Omelianchuk, Vitaliy Atrasevych, Artem Chernodub, and Oleksandr
  Skurzhanskyi. 2020.
\newblock Gector--grammatical error correction: Tag, not rewrite.
\newblock \emph{arXiv preprint arXiv:2005.12592}.

\bibitem[{Rao et~al.(2020)Rao, Yang, and Zhang}]{rao-etal-2020-overview}
Gaoqi Rao, Erhong Yang, and Baolin Zhang. 2020.
\newblock \href {https://aclanthology.org/2020.nlptea-1.4} {Overview of
  {NLPTEA}-2020 shared task for {C}hinese grammatical error diagnosis}.
\newblock In \emph{Proceedings of the 6th Workshop on Natural Language
  Processing Techniques for Educational Applications}, pages 25--35, Suzhou,
  China. Association for Computational Linguistics.

\bibitem[{Rothe et~al.(2021)Rothe, Mallinson, Malmi, Krause, and
  Severyn}]{rothe2021simple}
Sascha Rothe, Jonathan Mallinson, Eric Malmi, Sebastian Krause, and Aliaksei
  Severyn. 2021.
\newblock A simple recipe for multilingual grammatical error correction.
\newblock \emph{arXiv preprint arXiv:2106.03830}.

\bibitem[{Tseng et~al.(2015)Tseng, Lee, Chang, and
  Chen}]{tseng2015introduction}
Yuen-Hsien Tseng, Lung-Hao Lee, Li-Ping Chang, and Hsin-Hsi Chen. 2015.
\newblock Introduction to sighan 2015 bake-off for chinese spelling check.
\newblock In \emph{Proceedings of the Eighth SIGHAN Workshop on Chinese
  Language Processing}, pages 32--37.

\bibitem[{Vaswani et~al.(2017)Vaswani, Shazeer, Parmar, Uszkoreit, Jones,
  Gomez, Kaiser, and Polosukhin}]{vaswani2017attention}
Ashish Vaswani, Noam Shazeer, Niki Parmar, Jakob Uszkoreit, Llion Jones,
  Aidan~N Gomez, {\L}ukasz Kaiser, and Illia Polosukhin. 2017.
\newblock Attention is all you need.
\newblock In \emph{Advances in neural information processing systems}, pages
  5998--6008.

\bibitem[{Wang et~al.(2018{\natexlab{a}})Wang, Song, Li, Han, and
  Zhang}]{wang2018hybrid}
Dingmin Wang, Yan Song, Jing Li, Jialong Han, and Haisong Zhang.
  2018{\natexlab{a}}.
\newblock A hybrid approach to automatic corpus generation for chinese spelling
  check.
\newblock In \emph{Proceedings of the 2018 Conference on Empirical Methods in
  Natural Language Processing}, pages 2517--2527.

\bibitem[{Wang et~al.(2018{\natexlab{b}})Wang, Song, Li, Han, and
  Zhang}]{wang-etal-2018-hybrid}
Dingmin Wang, Yan Song, Jing Li, Jialong Han, and Haisong Zhang.
  2018{\natexlab{b}}.
\newblock \href {https://doi.org/10.18653/v1/D18-1273} {A hybrid approach to
  automatic corpus generation for {C}hinese spelling check}.
\newblock In \emph{Proceedings of the 2018 Conference on Empirical Methods in
  Natural Language Processing}, pages 2517--2527, Brussels, Belgium.
  Association for Computational Linguistics.

\bibitem[{Wang et~al.(2019)Wang, Tay, and Zhong}]{wang2019confusionset}
Dingmin Wang, Yi~Tay, and Li~Zhong. 2019.
\newblock Confusionset-guided pointer networks for chinese spelling check.
\newblock In \emph{Proceedings of the 57th Annual Meeting of the Association
  for Computational Linguistics}, pages 5780--5785.

\bibitem[{Wang et~al.(2020{\natexlab{a}})Wang, Kurosawa, Katsumata, and
  Komachi}]{wang-etal-2020-chinese}
Hongfei Wang, Michiki Kurosawa, Satoru Katsumata, and Mamoru Komachi.
  2020{\natexlab{a}}.
\newblock \href {https://aclanthology.org/2020.aacl-main.20} {{C}hinese
  grammatical correction using {BERT}-based pre-trained model}.
\newblock In \emph{Proceedings of the 1st Conference of the Asia-Pacific
  Chapter of the Association for Computational Linguistics and the 10th
  International Joint Conference on Natural Language Processing}, pages
  163--168, Suzhou, China. Association for Computational Linguistics.

\bibitem[{Wang et~al.(2020{\natexlab{b}})Wang, Wang, Gong, Wang, Hu, Duan,
  Shen, Yue, Fu, Wu, Che, Wang, Hu, and Liu}]{wang-etal-2020-combining-resnet}
Shaolei Wang, Baoxin Wang, Jiefu Gong, Zhongyuan Wang, Xiao Hu, Xingyi Duan,
  Zizhuo Shen, Gang Yue, Ruiji Fu, Dayong Wu, Wanxiang Che, Shijin Wang,
  Guoping Hu, and Ting Liu. 2020{\natexlab{b}}.
\newblock \href {https://aclanthology.org/2020.nlptea-1.5} {Combining
  {R}es{N}et and transformer for {C}hinese grammatical error diagnosis}.
\newblock In \emph{Proceedings of the 6th Workshop on Natural Language
  Processing Techniques for Educational Applications}, pages 36--43, Suzhou,
  China. Association for Computational Linguistics.

\bibitem[{Xu et~al.(2021)Xu, Li, Zhou, Li, Wang, Cao, Huang, and
  Mao}]{xu2021read}
Heng-Da Xu, Zhongli Li, Qingyu Zhou, Chao Li, Zizhen Wang, Yunbo Cao, Heyan
  Huang, and Xian-Ling Mao. 2021.
\newblock Read, listen, and see: Leveraging multimodal information helps
  chinese spell checking.
\newblock \emph{arXiv preprint arXiv:2105.12306}.

\bibitem[{Yu and Li(2014)}]{yu-li-2014-chinese}
Junjie Yu and Zhenghua Li. 2014.
\newblock \href {https://doi.org/10.3115/v1/W14-6835} {{C}hinese spelling error
  detection and correction based on language model, pronunciation, and shape}.
\newblock In \emph{Proceedings of The Third {CIPS}-{SIGHAN} Joint Conference on
  {C}hinese Language Processing}, pages 220--223, Wuhan, China. Association for
  Computational Linguistics.

\bibitem[{Yu et~al.(2014)Yu, Lee, Tseng, and Chen}]{yu2014overview}
Liang-Chih Yu, Lung-Hao Lee, Yuen-Hsien Tseng, and Hsin-Hsi Chen. 2014.
\newblock Overview of sighan 2014 bake-off for chinese spelling check.
\newblock In \emph{Proceedings of The Third CIPS-SIGHAN Joint Conference on
  Chinese Language Processing}, pages 126--132.

\bibitem[{Zhang et~al.(2021)Zhang, Pang, Zhang, Wang, He, Sun, Wu, and
  Wang}]{zhang2021correcting}
Ruiqing Zhang, Chao Pang, Chuanqiang Zhang, Shuohuan Wang, Zhongjun He, Yu~Sun,
  Hua Wu, and Haifeng Wang. 2021.
\newblock Correcting chinese spelling errors with phonetic pre-training.
\newblock In \emph{Findings of the Association for Computational Linguistics:
  ACL-IJCNLP 2021}, pages 2250--2261.

\bibitem[{Zhang et~al.(2020)Zhang, Huang, Liu, and
  Li}]{zhang-etal-2020-spelling}
Shaohua Zhang, Haoran Huang, Jicong Liu, and Hang Li. 2020.
\newblock \href {https://doi.org/10.18653/v1/2020.acl-main.82} {Spelling error
  correction with soft-masked {BERT}}.
\newblock In \emph{Proceedings of the 58th Annual Meeting of the Association
  for Computational Linguistics}, pages 882--890, Online. Association for
  Computational Linguistics.

\bibitem[{Zhang et~al.(2015)Zhang, Xiong, Hou, Zhang, and
  Cheng}]{zhang-etal-2015-hanspeller}
Shuiyuan Zhang, Jinhua Xiong, Jianpeng Hou, Qiao Zhang, and Xueqi Cheng. 2015.
\newblock \href {https://doi.org/10.18653/v1/W15-3107} {{HANS}peller++: A
  unified framework for {C}hinese spelling correction}.
\newblock In \emph{Proceedings of the Eighth {SIGHAN} Workshop on {C}hinese
  Language Processing}, pages 38--45, Beijing, China. Association for
  Computational Linguistics.

\bibitem[{Zhang et~al.(2019)Zhang, Feng, Meng, You, and
  Liu}]{zhang-etal-2019-bridging}
Wen Zhang, Yang Feng, Fandong Meng, Di~You, and Qun Liu. 2019.
\newblock \href {https://doi.org/10.18653/v1/P19-1426} {Bridging the gap
  between training and inference for neural machine translation}.
\newblock In \emph{Proceedings of the 57th Annual Meeting of the Association
  for Computational Linguistics}, pages 4334--4343, Florence, Italy.
  Association for Computational Linguistics.

\bibitem[{Zhao et~al.(2018)Zhao, Jiang, Sun, and Wan}]{zhao2018overview}
Yuanyuan Zhao, Nan Jiang, Weiwei Sun, and Xiaojun Wan. 2018.
\newblock Overview of the nlpcc 2018 shared task: Grammatical error correction.
\newblock In \emph{CCF International Conference on Natural Language Processing
  and Chinese Computing}, pages 439--445. Springer.

\end{thebibliography}
\bibliographystyle{acl_natbib}

\appendix
\section{Threshold Analysis}
\begin{table*}[htbp]
\small
\centering
\begin{tabular}{l|cccc|cccc}
\hline
\multirow{2}{*}{Method}                                                   & \multicolumn{1}{l}{\multirow{2}{*}{Threshold}} & \multicolumn{3}{c|}{Insertion-dev} & \multicolumn{1}{l}{\multirow{2}{*}{Threshold}} & \multicolumn{3}{c}{Deletion-dev} \\
                                                                          & \multicolumn{1}{l}{}                           & Precision    & Recall    & F1     & \multicolumn{1}{l}{}                           & Precision      & Recall  & F1    \\ \hline
\multirow{6}{*}{BERT w/ substituted \texttt{[mask]}} & 1e-4                                           & 19.6         & 9.2       & 12.6   & 1e-4                                           & 21.9           & 12.1    & 15.6  \\
                                                                          & 5e-4                                           & 16.3         & 14.7      & 15.5   & 3e-4                                           & 19.3           & 17.6    & 18.4  \\
                                                                          & 1e-3                                           & 14.9         & 17.4      & 16.1   & 5e-4                                 & 18.6  &19.9    & 19.3  \\
                                                                          & \textbf{3e-3}                                           & \textbf{13.1}         & \textbf{24.0}      & \textbf{16.9}   & \textbf{1e-3 }                                          & \textbf{18.6}           & \textbf{26.1}    & \textbf{21.7}  \\
                                                                          & 5e-3                                           & 12.0         & 24.9      & 16.2   & 3e-3                                           & 15.6           & 34.3    & 21.4  \\
                                                                          & 1e-2                                           & 11.1         & 28.4      & 15.9   & 5e-3                                           & 14.8           & 37.6    & 21.2  \\ \hline
\multirow{6}{*}{BERT w/o \texttt{[mask]}}            & 0.8                                            & 21.8         & 6.2       & 9.7    & 0.3                                            & 44.4           & 8.4     & 14.2  \\
                                                                          & 0.9                                            & 20.5         & 8.3       & 11.8   & 0.5                                            & 38.6           & 10.3    & 16.3  \\
                                                                          & 0.98                                           & 17.2         & 14.4      & 15.7   & 0.7                                            & 35.8           & 11.9    & 17.8  \\
                                                                          & 0.992                                          & 14.2         & 17.7      & 15.7   & 0.9                                            & 34.7           & 18.4    & 24.0  \\
                                                                          & \textbf{0.994}                                          & \textbf{14.0}         & \textbf{20.0}      & \textbf{16.4}   & \textbf{0.99}                                           & \textbf{22.8}           & \textbf{34.5}    & \textbf{27.5}  \\
                                                                          & 0.996                                          & 13.1         & 22.1      & 16.4   & 0.992                                          & 21.8           & 36.6    & 27.3  \\ \hline
\multirow{6}{*}{BERT w/ inserted \texttt{[mask]}}    & 0.8                                            & 5.0          & 42.5      & 8.9    & -                                              & -              & -       & -     \\
                                                                          & 0.9                                            & 7.0          & 46.0      & 12.2   & -                                              & -              & -       & -     \\
                                                                          & 0.95                                           & 9.2          & 43.0      & 15.2   & -                                              & -              & -       & -     \\
                                                                          & 0.99                                           & 17.4         & 34.8      & 23.2   & -                                              & -              & -       & -     \\
                                                                          & \textbf{0.998}                                          & \textbf{30.9}         & \textbf{23.8}      & \textbf{26.9}   & -                                              & -              & -       & -     \\
                                                                          & 0.999                                          & 36.9         & 18.9      & 25.3   & -                                              & -              & -       & -     \\ \hline 
\multirow{6}{*}{MacBERT}                                                  & 0.01                                           & 6.4          & 11.8      & 8.3    & 0.01                                           & 11.2           & 28.2    & 16.0  \\
                                                                          & 0.03                                           & 6.7          & 16        & 9.5    & \textbf{0.02}                                           & \textbf{11.3}           & \textbf{33.9}    & \textbf{17.0}  \\
                                                                          & 0.05                                           & 7.1          & 19.1      & 10.4   & 0.03                                           & 10.9           & 35.4    & 16.7  \\
                                                                          & 0.07                                           & 6.9          & 20.1      & 10.2   & 0.05                                           & 103            & 37.9    & 16.2  \\
                                                                          & \textbf{0.1 }                                           & \textbf{6.7}          & \textbf{22.4}      & \textbf{10.4}   & 0.1                                            & 9.4            & 43.1    & 15.5  \\
                                                                          & 0.2                                            & 5.7          & 25.6      & 9.4    & 0.2                                            & 8.1            & 49.8    & 13.9  \\ \hline 
\multirow{6}{*}{PLOME}                                                    & 0.7                                            & 7.0          & 15.3      & 9.6    & 0.5                                            & 10.8           & 25.0    & 15.1  \\
                                                                          & 0.8                                            & 7.0          & 17.7      & 10.0   & 0.6                                            & 10.7           & 27.4    & 15.4  \\
                                                                          & 0.9                                            & 6.8          & 21.6      & 10.3   & 0.7                                            & 10.4           & 29.8    & 15.4  \\
                                                                          & 0.95                                           & 6.6          & 26.2      & 10.5   & 0.8                                            & 10.3           & 34.0    & 15.8  \\
                                                                          & \textbf{0.98}                                           & \textbf{6.4}          & \textbf{34.2 }     & \textbf{ 10.8}   & \textbf{0.85}                                           & \textbf{10.2}           &\textbf{ 37.2}    & \textbf{16.0}  \\
                                                                          & 0.99                                           & 6.0          & 40.2      & 10.5   & 0.9                                            & 9.9            & 41.5    & 16.0  \\ \hline
\multirow{6}{*}{Our Approach}                                             & 0.06                                           &  77.0            & 73.2          & 75.0       & 0.8                                            & 33.2           & 92.2    & 48.8  \\
                                                                          & \textbf{0.07}                                           & \textbf{76.2}         & \textbf{75.2}      & \textbf{75.7}   & 0.9                                            & 41.8           & 88.9    & 56.8  \\
                                                                          & 0.08                                           & 74.9         & 76.3      & 75.6   & 0.98                                           & 59.9           & 78.4    & 67.9  \\
                                                                          & 0.09                                           & 73.8         & 76.7      & 75.2   & 0.99                                           & 66.3           & 72.0    & 69.1  \\
                                                                          & 0.1                                            & 72.7         & 77.6      & 75.1   & \textbf{0.991}                                          & \textbf{67.3}           &\textbf{ 71.1 }   & \textbf{69.3 } \\
                                                                          & 0.12                                           & 70.1         & 79.0      & 74.3   & 0.992                                          & 68.3           & 69.4    & 68.8 \\ \hline
\end{tabular}
\caption{The thresholds that selected for each method.}
\label{table:threshold}
\end{table*}

\begin{figure*}
\begin{subfigure}{.5\textwidth}
  \centering
  \includegraphics[width=0.98\linewidth]{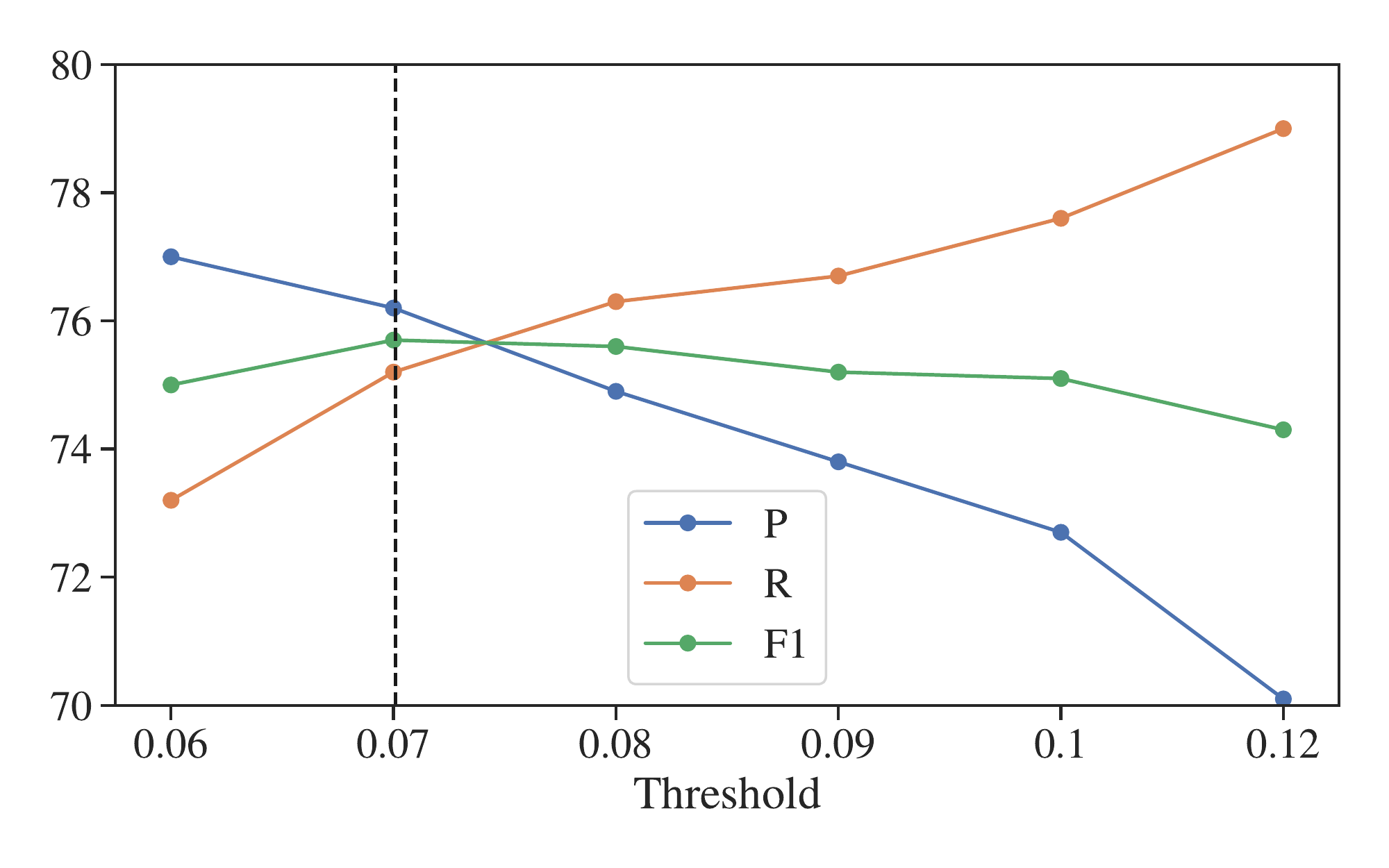}
  \caption{Word Insertion}
  \label{fig:insert-threshold}
\end{subfigure}%
\begin{subfigure}{.5\textwidth}
  \centering
  \includegraphics[width=0.97\linewidth]{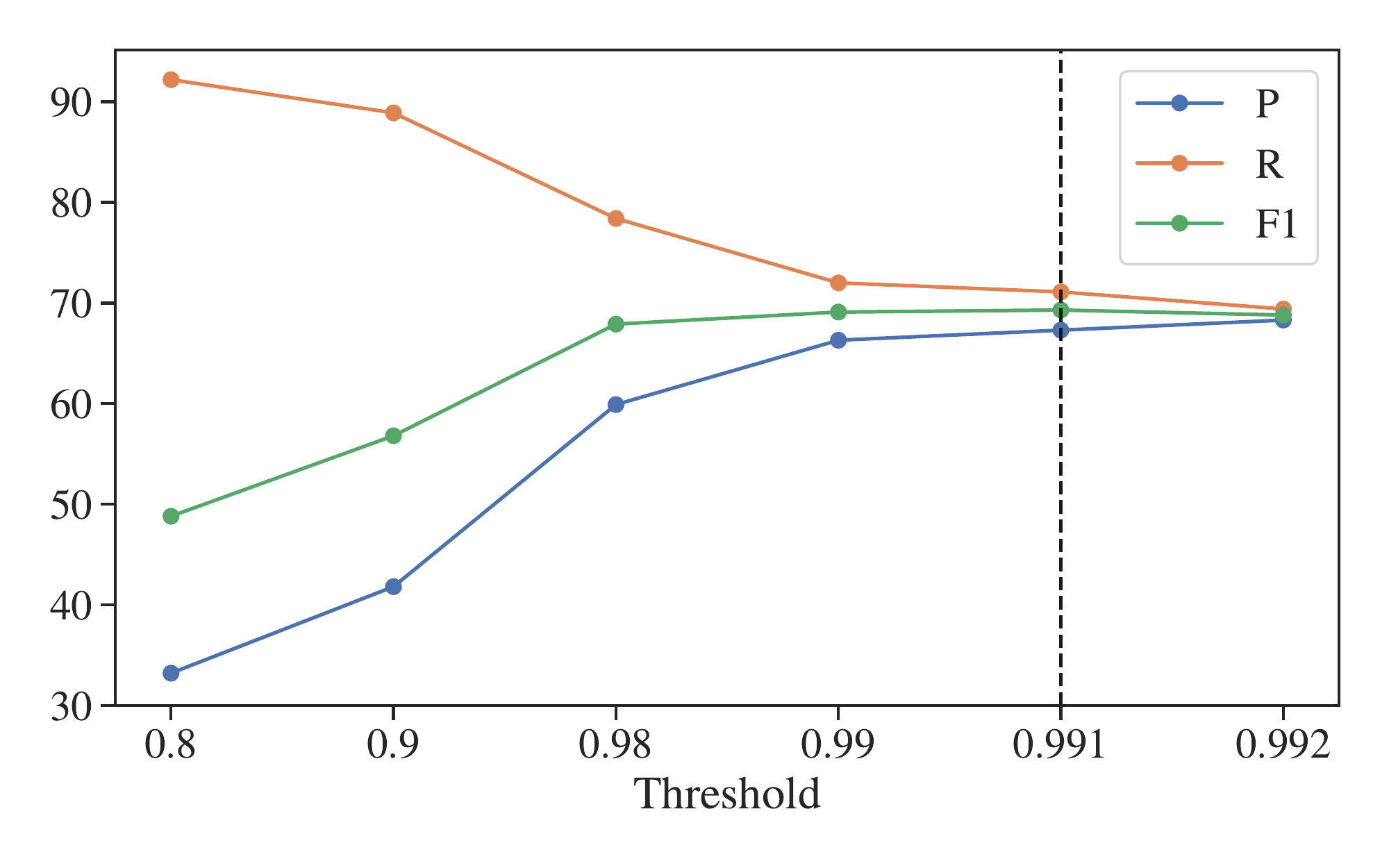}
  \caption{Word Deletion}
  \label{fig:delete-threshold}
\end{subfigure}
\caption{The performance of our approach on detecting deletion and insertion errors with different thresholds. Numbers are reported on the dev set. The finally selected thresholds are marked with dot lines.}
\label{fig:threshold}
\end{figure*}

\begin{table*}[t]
\centering
\begin{tabular}{l|ccc|ccc}
\hline
\multicolumn{1}{c|}{\multirow{2}{*}{Method}} & \multicolumn{3}{c|}{Insertion}                 & \multicolumn{3}{c}{Deletion} \\ \cline{2-7} 
\multicolumn{1}{c|}{}                        & P & R & F1   & P   & R  & F1    \\ \hline
\multicolumn{1}{l|}{BERT w/ substituted \texttt{[mask]}}               & 12.2      & 0.8   & \multicolumn{1}{l|}{1.5} & 55.9        & 3.3    & 6.2  \\
\multicolumn{1}{l|}{BERT w/o \texttt{[mask]}}            & 8.9      & 19.7    & \multicolumn{1}{l|}{12.2} & 27.4       & 29.0    & 28.2  \\
\multicolumn{1}{l|}{BERT w/ inserted \texttt{[mask]}}             & 12.6          & 11.6       & \multicolumn{1}{l|}{11.8}     &  -           &   -      &    -   \\
\multicolumn{1}{l|}{Flying}                   & 35.3          & 11.4       & \multicolumn{1}{l|}{17.2}     & 40.9           &  13.9       &  20.8     \\ 
\multicolumn{1}{l|}{Our approach}              & \textbf{43.7}      & \textbf{39.5}   & \multicolumn{1}{l|}{\textbf{41.5}} & \textbf{41.2}        & \textbf{39.7}    & \textbf{40.4}  \\ \hline
\end{tabular}
\caption{Results on CGED.}
\label{table:results-detection}
\end{table*}


\end{document}